\pdfoutput=1
\documentclass[11pt]{article}

\usepackage[]{ACL2023}

\usepackage{times}
\usepackage{latexsym}
\usepackage{comment}
\usepackage[T1]{fontenc}
\usepackage[utf8]{inputenc}
\usepackage{todonotes}
\usepackage{microtype}
\usepackage{inconsolata}
\usepackage[english]{babel}
\usepackage[dvips]{epsfig}
\usepackage{amsmath}
\usepackage{amssymb}
\usepackage{amsfonts}
\usepackage{amsthm}
\usepackage{amsbsy}
\usepackage{amsgen}
\usepackage{amscd}
\usepackage{amsopn}
\usepackage{amstext}
\usepackage{amsxtra}
\usepackage{mathrsfs}
\usepackage{enumitem}
\usepackage{graphicx}
\usepackage{verbatim}
\usepackage{float}
\usepackage[all,cmtip]{xy}
\usepackage{accents}
\usepackage{sseq}
\usepackage{url}
\usepackage{hyperref}
\usepackage{makeidx}
\usepackage{xspace}

\usepackage{tabularx}
\usepackage{array}
\usepackage{multirow}
\usepackage{booktabs}
\usepackage{anyfontsize}
\usepackage{multirow}
\usepackage{makecell}

\usepackage{listings}
\definecolor{codegreen}{rgb}{0,0.6,0}
\definecolor{codegray}{rgb}{0.5,0.5,0.5}
\definecolor{codepurple}{rgb}{0.58,0,0.82}
\definecolor{backcolour}{rgb}{0.95,0.95,0.92}
\lstdefinestyle{mystyle}{
    backgroundcolor=\color{backcolour},   
    commentstyle=\color{codegreen},
    keywordstyle=\color{magenta},
    numberstyle=\tiny\color{codegray},
    stringstyle=\color{codepurple},
    basicstyle=\ttfamily\footnotesize,
    breakatwhitespace=false,         
    breaklines=true,                 
    captionpos=b,                    
    keepspaces=true,                 
    numbers=left,                    
    numbersep=5pt,                  
    showspaces=false,                
    showstringspaces=false,
    showtabs=false,                  
    tabsize=2
}

\newenvironment{itemize*}%
  {\begin{itemize}%
    \setlength{\itemsep}{0.9pt}%
    \setlength{\parskip}{0.9pt}%
    \setlength{\topsep}{0.9pt}}%
  {\end{itemize}}

\newcommand{\rst}{ROOTS Search Tool\xspace}
\newcommand{\hf}{$^1$}
\newcommand{\sapienza}{$^2$}
\newcommand{\hfsapienza}{$^{1,2}$}
\newcommand{\lu}{$^3$}
\newcommand{\scad}{$^4$}
\newcommand{\luscad}{$^{3,4}$}
\newcommand{\telefonica}{$^5$}
\newcommand{\mavenoid}{$^6$}
\newcommand{\uc}{$^7$}

\title{The \rst: Data Transparency for LLMs}

\author{
Aleksandra Piktus\hfsapienza{}
Christopher Akiki\luscad{} 
Paulo Villegas\telefonica{}
Hugo Laurençon\hf{}\\
\textbf{
Gérard Dupont\mavenoid{}
Alexandra Sasha Luccioni\hf{}
Yacine Jernite\hf{}
Anna Rogers\uc {}
} \\
\\
\hf{}Hugging Face \ \sapienza{}Sapienza University \ \lu{}Leipzig University \ \scad{}ScaDS.AI \\ 
\telefonica{}Telefonica I+D \mavenoid{}Mavenoid \ \uc{}University of Copenhagen \\
{\tt piktus@huggingface.co} \\
\\
}

\date{}

\begin{document}
\maketitle

\begin{abstract}
ROOTS is a 1.6TB multilingual text corpus developed for the training of BLOOM, currently the largest language model explicitly accompanied by commensurate data governance efforts. In continuation of these efforts, we present the \rst: a search engine over the entire ROOTS corpus offering both fuzzy and exact search capabilities.
ROOTS is the largest corpus to date that can be investigated this way. The \rst is open-sourced and available \href{https://huggingface.co/spaces/bigscience-data/roots-search}{on Hugging Face Spaces}. We describe our implementation and the possible use cases of our tool.
\end{abstract}

\section{Introduction}
Large language models (LLMs) are ubiquitous in modern NLP, used directly to generate text and as building blocks in downstream applications. The ever-increasing size of the latest models inflates the demand for massive volumes of training data~\cite{https://doi.org/10.48550/arxiv.2203.15556}, in practice sourced mainly from the Web. This raises questions concerning the quality of the data, the feasibility of curating and inspecting it, as well as documenting it in terms of what kinds of speech and speakers it represents \cite{Jo_2020,BenderGebruEtAl_2021_On_Dangers_of_Stochastic_Parrots_Can_Language_Models_Be_Too_Big,akiki2022bigscience}. Without that level of characterization, we cannot tell for what varieties of language the resulting models can be expected to work well, whether the data was ethically sourced, how to interpret evaluation metrics, and to what degree a particular output was memorized directly from the training data.
In an encouraging new trend, we see researchers exploring ways to quantitatively describe large datasets~\cite{mitchell-measuring-data-22}. However, user-friendly tools for an extensive qualitative analysis are still predominantly missing. In our current work, we aim to fill that gap for a specific, web-scale, textual corpus.

Building on the efforts of the BigScience workshop,\footnote{\url{bigscience.huggingface.co}} we present the \rst\footnote{\url{hf.co/spaces/bigscience-data/roots-search}}---a search engine for the the 1.6TB multilingual ROOTS corpus \cite{laurencon2022the}. The ROOTS corpus was created to pre-train BLOOM \cite{bloom}---the first LLM of its scale designed with commensurate efforts in responsible licensing\footnote{\url{bigscience.huggingface.co/blog/the-bigscience-rail-license}} and data governance \cite{10.1145/3531146.3534637}. We hope that our tool will facilitate qualitative analysis of the web-scale ROOTS corpus, and establish the qualitative analysis of training data---critical for the model understanding and governance work---as an essential step in the development of LLMs.

\begin{figure}[!t]

\includegraphics[width=\linewidth]{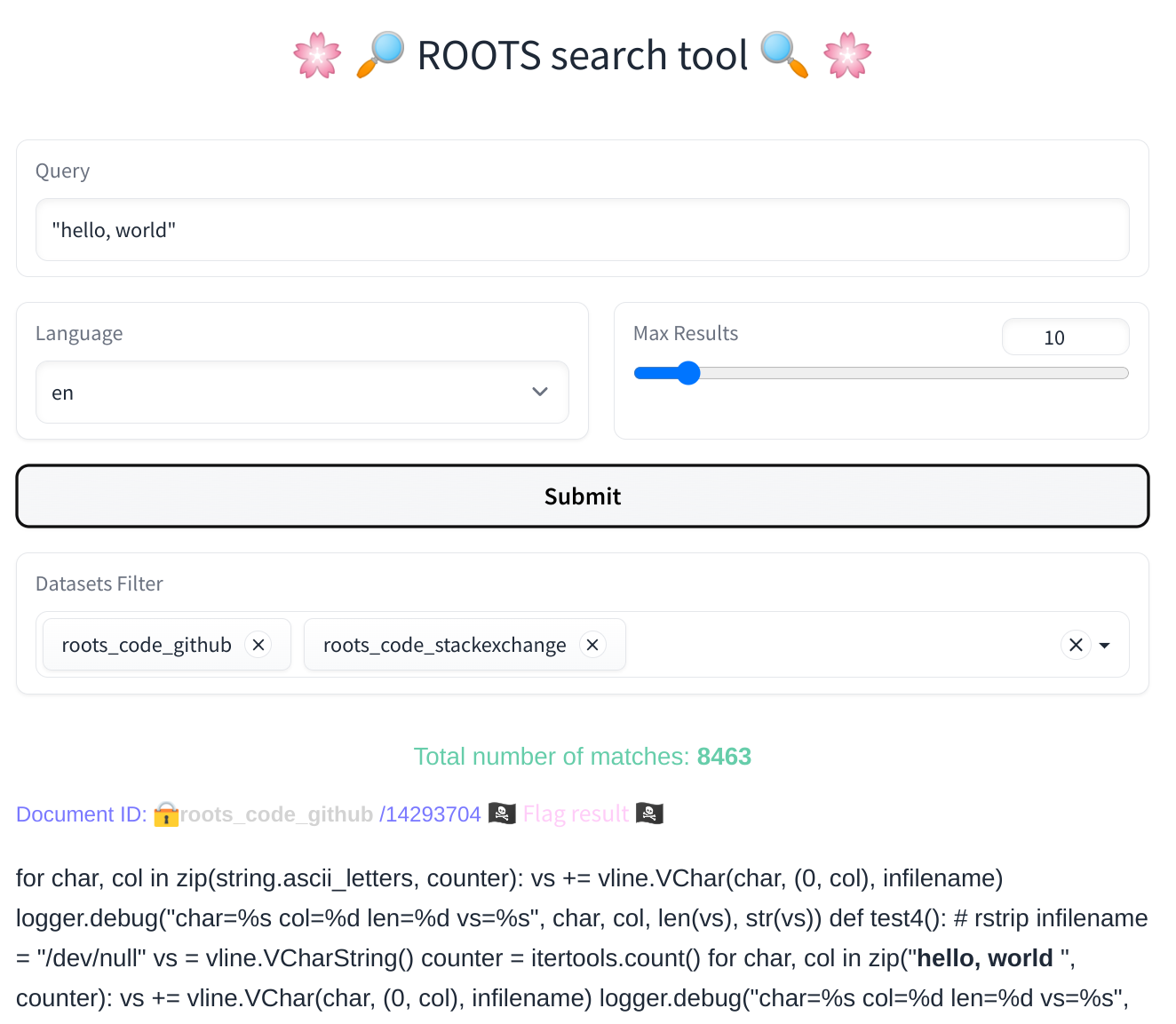}

\caption{ROOTS search tool: user interface}
\label{fig:ui}
\end{figure}

\section{Related Work}

\paragraph{Corpus linguistics.} The core methodology for studying large volumes of text was developed in corpus linguistics \cite{McEneryHardie_2013_History_of_Corpus_Linguistics}, an area of research responsible for curating large text collections carefully designed to represent specific varieties of language. For example, the 100M word British National Corpus~\cite{Leech_1992_100_million_words_of_English_British_National_Corpus_BNC} was created to represent the spoken and written British English of the late 20th century, with each text handpicked by experts, who also procured appropriate copyright exemptions. Similar national corpora were later created for many other languages, e.g. Japanese~\cite{Maekawa_2008_Balanced_Corpus_of_Contemporary_Written_Japanese}. The texts were often accompanied by multiple layers of annotations---syntactic, morphological, semantic, genre, source etc. This enabled valuable empirical research on the variants of represented languages, finding use in early distributional semantic models. Corpus linguistics developed sophisticated methodologies including concordances, word sketches and various word association measures \cite[][among others]{StefanowitschGries_2003_Collostructions_Investigating_interaction_of_words_and_constructions,Baker_2004_Querying_Keywords_Questions_of_Difference_Frequency_and_Sense_in_Keywords_Analysis,Kilgarriff_2014_Sketch_Engine_ten_years_on}. However, this methodology did not adapt well to Web-scale corpora due to the lack of tools and resources that could support such scale.

\paragraph{Web-scale corpora for LLM pre-training.}
As LLMs grew, so did the need for massive pre-training datasets. To date, there were several efforts to collect and clean large English and multilingual corpora \cite{JMLR:v21:20-074,XueConstantEtAl_2021_mT5_Massively_Multilingual_Pretrained_TexttoText_Transformer,GaoBidermanEtAl_2020_Pile_800GB_Dataset_of_Diverse_Text_for_Language_Modeling,OrtizSuarezRomaryEtAl_2020_Monolingual_Approach_to_Contextualized_Word_Embeddings_for_MidResource_Languages,BanonChenEtAl_2020_ParaCrawl_WebScale_Acquisition_of_Parallel_Corpora,El-KishkyChaudharyEtAl_2020_CCAligned_Massive_Collection_of_CrossLingual_WebDocument_Pairs}. Non-English, monolingual corpora of this scale have also started to emerge~\cite{Gutierrez-FandinoPerez-FernandezEtAl_2022_esCorpius_Massive_Spanish_Crawling_Corpus,KummervoldWetjenEtAl_2022_Norwegian_Colossal_Corpus_Text_Corpus_for_Training_Large_Norwegian_Language_Models}
However, the sheer scale of such datasets renders them hard to properly curate: we now know that the data used for training LLMs may contain synthetic data \cite{DodgeSapEtAl_2021_Documenting_Large_Webtext_Corpora_Case_Study_on_Colossal_Clean_Crawled_Corpus}, privacy-infringing data \cite{CarliniTramerEtAl_2020_Extracting_Training_Data_from_Large_Language_Models,HuangShaoEtAl_2022_Are_Large_PreTrained_Language_Models_Leaking_Your_Personal_Information}, incorrect language codes or and translations \cite{KreutzerCaswellEtAl_2022_Quality_at_Glance_Audit_of_WebCrawled_Multilingual_Datasets}, not to mention the ubiquitous issues with social biases \cite[][among others]{BlodgettBarocasEtAl_2020_Language_Technology_is_Power_Critical_Survey_of_Bias_in_NLP,FieldBlodgettEtAl_2021_Survey_of_Race_Racism_and_AntiRacism_in_NLP,StanczakAugenstein_2021_Survey_on_Gender_Bias_in_Natural_Language_Processing}. Another issue pertains to the permissions to use the data, which, perhaps the most famously, surfaced in relation to the BookCorpus \cite{ZhuKirosEtAl_2015_Aligning_Books_and_Movies_Towards_Story-Like_Visual_Explanations_by_Watching_Movies_and_Reading_Books}, used, among others, to train BERT~\cite{devlin-etal-2019-bert}, but collected without author permissions and eventually taken down by the authors \cite{https://doi.org/10.48550/arxiv.2105.05241}. 

These issues are a consequence of the fact that the current web-scale corpora are opportunistic samples of publicly available text, rather than artifacts curated to provide a representative snapshot of a specific language variety, as in the corpus linguistics work \cite{Rogers_2021_Changing_World_by_Changing_Data}. This highlights the general problem with the lack of documentation in NLP datasets of all sizes~\cite{BenderFriedman_2018_Data_Statements_for_Natural_Language_Processing_Toward_Mitigating_System_Bias_and_Enabling_Better_Science,GebruMorgensternEtAl_2020_Datasheets_for_Datasets}, and the fact that data work has generally not been a priority in NLP recently~\cite{SambasivanKapaniaEtAl_2021_Everyone_wants_to_do_model_work_not_data_work_Data_Cascades_in_High-Stakes_AI}.

\paragraph{Information Retrieval for massive text corpora.}
Inspecting large data collection is a central topic of study in another Machine Learning domain, namely Information Retrieval. Even though multiple techniques for analysing large document collections have been developed over the years, there has been little interest so far in applying them specifically to study LLM training data. The closest to out work is the C4~\cite{JMLR:v21:20-074} Search \footnote{\url{https://c4-search.apps.allenai.org/}}, however, the tool comes with no documentation to explain the details of the indexed variant of the dataset or applied design choices. Similar tools emerge for smaller, more specialised corpora, e.g. COVID-related datasets~\citep{zhang2020covidex}, news quotes~\cite{Vukovi__2022} and medical literature~\citep{cancer-nlp-no-code}.
\citet{RazeghiMekalaEtAl_2022_Snoopy_Online_Interface_for_Exploring_Effect_of_Pretraining_Term_Frequencies_on_FewShot_LM_Performance} provide an interface to pre-computed term frequencies from the Pile, but it does not provide full-text corpus search. In the Computer Vision community, related efforts\footnote{\url{https://haveibeentrained.com/}} target large text and image datasets such as LAION~\cite{SchuhmannBeaumontEtAl_2022_LAION5B_open_largescale_dataset_for_training_next_generation_imagetext_models,SchuhmannVencuEtAl_2021_LAION400M_Open_Dataset_of_CLIPFiltered_400_Million_ImageText_Pairs}.

 We believe our work to be the first principled effort in providing search access to the training corpus of an existing large language model and the largest text dataset search tool currently available.

\section{The ROOTS corpus}
The ROOTS corpus~\cite{laurencon2022the} is a high-quality, heterogeneous, multilingual text corpus collected as part of the BigScience project to train the BLOOM~LLM~\cite{bloom}. ROOTS consists of 1.6TB of data in 46 natural and 13 programming languages. The full ROOTS dataset is open to the members of the \href{https://hf.co/bigscience-data}{BigScience Data organization} on the Hugging Face hub, which the interested researchers can still apply to join\footnote{Sign-up link is available \href{https://hf.co/spaces/bigscience-data/roots-search}{here}}.

\subsection{Data Governance}
The development of the BLOOM model within the BigScience project was backed by significant work on data governance, as it is was identified early on as one of the highest-impact levers of action to enable better accountability and data subject agency in modern ML technology\footnote{\href{https://montrealethics.ai/social-context-of-llms-the-bigscience-approach-part-3-data-governance-and-representation/}{Data governance and representation in BigScience.}}. Participants started by designing a new governance framework to meet the unique needs of distributed data governance for web-scale data in terms of respecting data subject rights~\cite{10.1145/3531146.3534637}. A partial implementation of this framework was used for the ROOTS data as described by \citet{laurencon2022the}, focusing on explicit agreements with data custodians, extensive documentation of the data sources, technical tools for privacy-enhancing data handling, and purpose-specific access to subsets of the data.

The present tool goes one step further in implementing the proposed data governance feedback by enabling examination and feedback for the data sources from any interested parties; while still maintaining the controlled access necessary to the proposed governance. The tool only provides 128-word snippets of indexed documents, akin to regular web search engines, and hence provides no practical way to reconstruct the full corpus. The snippets are traceable to their origin in the full ROOTS corpus, and we additionally link to original source documents whenever possible.\footnote{Unfortunately, the metadata in ROOTS is inconsistent and we only have access to URLs in the \texttt{pseudocrawl} datasets.} Finally, users of the tool are able to flag specific search results with an explanation to outline possible infringements of data subjects' privacy or intellectual property rights. At this stage, the information collected from the flagging process is primarily intended to serve as a basis for  future research on collaborative data governance processes. We provide more examples of use cases to support data examination and governance in Section~\ref{sec:use-cases}.

\subsection{Data Pre-processing}
\paragraph{Documents vs snippets.} ROOTS consists of documents of varying lengths, with outliers as long as 282,571 words. For fuzzy search, we split documents into short snippets of at most 128 words and index snippets rather than the original documents. This helps us follow the controlled access principle discussed in the previous section and makes indexed snippets more comparable in the context of fuzzy search. In exact search, we look for the exact occurrences of the input query within documents and construct snippets ad hoc, including words on both sides of the detected occurrence.

\paragraph{Unique Result IDs.} 
In order to be able to trace search results back to their source, we construct result IDs, adopting the following convention: (a)~we include the dataset name as defined on the Hugging Face Hub, followed by (b)~the ID of the document from which the given snippet came, (c)~and a question mark. We then include parameters which differ depending on the search strategy used. In fuzzy search we introduce two parameters: the \texttt{seg} parameter describing the segmentation strategy applied during the pre-processing stage, and the \texttt{seg\_id} parameter indicating the rank of the given snippet under the specified segmentation strategy. For exact search, we include a single \texttt{id} parameter indicating the the rank of the occurrence of the query in the current document.

\paragraph{PII redaction.}
\label{sec:PII}
\begin{figure}[!t]

\includegraphics[width=\linewidth]{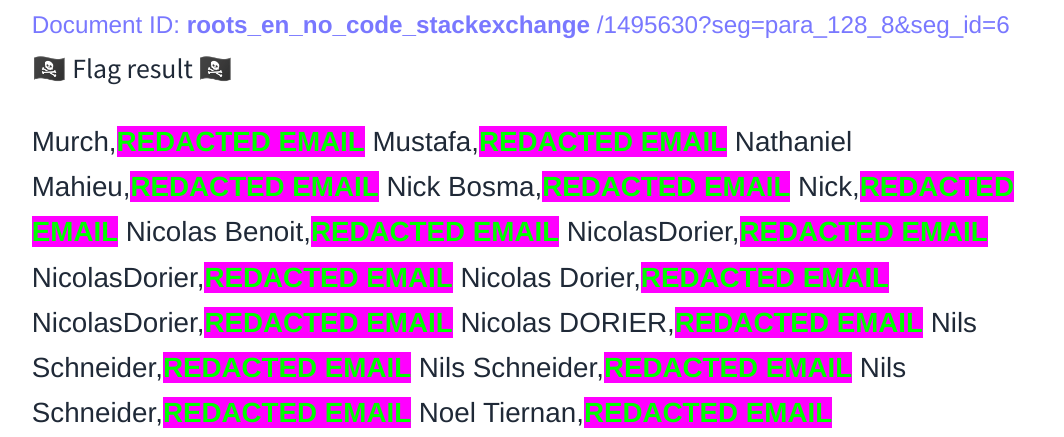}

\caption{PII leakage: example result for the query \texttt{gmail.com}. We indicate the redacted PII with green and pink treatment.}
\label{fig:pii_leakage}
\end{figure}

During preliminary experiments on the ROOTS corpus, OSCAR~\cite{OrtizSuarezSagotRomary2019} has been identified as a source of a large amount of documents containing personally identifiable information (PII). A regular-expression-based PII redaction script\footnote{The BigScience PII redaction script is available~\href{https://github.com/bigscience-workshop/data-preparation/tree/main/preprocessing/training/02_pii}{here}} has been applied to OSCAR prior to BLOOM training. However, the dataset itself still contains unredacted text. In order to avoid leaking PII through our search tool, we apply an improved variant of the BigScience PII redaction script on the backend side and display results with PII redacted in a visible way - this way one can inspect the data and observe the problem, but personal information are predominantly removed. An example is shown in \autoref{fig:pii_leakage}.

\begin{table*}[t!]
\footnotesize
\centering
\begin{tabular}{rrrrrc}
\toprule
 \makecell[r]{ROOTS language tag} & \# documents & \makecell[r]{Data size (GB)} & \# snippets & \makecell[r]{Index size (GB)} & Analyzer\\
 \midrule
zh, zhs, zht		& 88,814,841	&  259.01 & 111,284,681	& 682 & zh\\
indic	& 84,982,982 & 70.45	& 100,810,124	& 714.08 & whitespace \\
en	& 77,010,827& 470.47	& 695,521,432	& 766.14 & en\\
es	& 67,005,817	& 172.40	& 267,542,136	& 264.35 & es\\
fr	& 58,847,091& 204.03	& 299,938,546	& 305.29 & fr\\
vi		& 34,110,375	& 42.83 & 76,164,552	& 72.89 & whitespace \\
pt	& 31,969,891& 77.59	& 122,221,863	& 119.98 & pt \\
code	& 26,176,998& 173.16	& 365,424,222	& 206.96 & whitespace \\
ar	& 15,234,080	& 73.75	& 68,509,441	& 93.71 & ar \\
id		& 12,514,253	& 19.63 & 29,531,873	& 27.16 & id\\
ca		& 6,142,390	& 17.42 & 26,844,600	& 29.65 & es \\
eu		& 5,149,797	& 2.36 & 6,219,039	& 4.56 & whitespace\\
nigercongo	& 1,162,568	& 0.48	& 1,462,238	& 0.89 & whitespace\\
 \midrule
total	&	597,936,751	 & 1583.59 & 2,171,474,747	& 2518.99 \\
\bottomrule
\end{tabular}
\caption{Each row represents a single BM25 index we build.}
\label{tab:data_sizes}
\end{table*}

\section{Implementation}
\paragraph{Fuzzy Search Backend.}
The ROOTS corpus is organized in 498~datasets, each annotated with a language identifier. There are two types of identifiers: those indicating an individual language (e.g. \texttt{pt} for Portuguese), and those indicating a language within a language group (e.g. \texttt{indic-mr} for Marathi, as part of the Indic language group). All programming languages are collected under a common \texttt{code} tag. We build 13 sparse, BM25~\cite{robertson2009probabilistic} indices: one per language group for the \texttt{indic} and \texttt{nigercongo} groups, one for \texttt{code}, and one for each of the remaining languages (except Chinese, where we combine the tags \texttt{zh}, \texttt{zht}, and \texttt{zhs} into a single index).
Table~\ref{tab:data_sizes} presents the basic information per index.
We index respective subsets of the corpus using Pyserini~\cite{Lin_etal_SIGIR2021_Pyserini}, a leading toolkit for reproducible IR research. Tokenization is performed with native Lucene\footnote{\url{https://lucene.apache.org/}} analyzers available via Pyserini API (see Table~\ref{tab:data_sizes} to check which analyzers were used for specific indices).

\paragraph{Exact Search Backend.}
We leverage a suffix array implementation\footnote{\url{https://github.com/google-research/deduplicate-text-datasets}} proposed by~\citet{lee2021deduplicating}. We build the suffix array for the whole ROOTS corpus, this time without the split into languages or language groups. We host both the BM25 indices and the suffix array on Hugging Face-provisioned machines. The server code is open-sourced\footnote{\url{https://github.com/huggingface/roots-search-tool}}.

\paragraph{Frontend and User Experience.}
The \rst user interface is built with Gradio~\cite{abid-2019-gradio} and served via Hugging Face Spaces.\footnote{\url{https://huggingface.co/docs/hub/spaces}}.
By default, searches are performed in fuzzy mode, in order to move to the exact search one can enclose the query in double quotes.
Fuzzy searches can be performed in a user-specified language, or in \textit{all} languages (in that case results are surfaced separately for each language). We also provide an option to auto-detect the language of the query with a FastText~\cite{joulin-etal-2017-bag} classifier. Results are displayed in the order of decreasing relevance; users can control the maximum number of results they want to see using a slider. In exact search mode, the backend returns all documents matching a given query exactly, irrespective of the language, and they are displayed over multiple pages in a random order, with the max results parameter controlling the size of a single page. The total number of matched results is displayed at the top of the results page. PII redaction is applied to all results on the backend side. The tool also allows users to filter out all results from a specific dataset appearing on a given page.
\section{Use cases}
\label{sec:use-cases}

\paragraph{Detecting PII issues to improve obfuscation.}
BLOOM was trained with efforts to detect and obfuscate PII in the original ROOTS documents, and as described in \autoref{sec:PII}, we build on that effort when obfuscating PII in search results. However, it is still possible that some such data was not detected. The tool allows searching for the specific PII by concerned individuals, which is the first step for requesting removal of their data. One could also simply search for their name to see if they are represented in the corpus, and how.

\paragraph{Detecting problematic content.}
Text from Web crawls is not necessarily high-quality human-written text. Among the possible problems are hate speech, excessive pornography, synthetic text (e.g. machine-translated text, AI-generated text), word lists that are not meaningful and are meant to trick search engines~\citep{Hamilton2013GooglebombingM}, factually incorrect text such as fake news or conspiracy theories. For example, we found at least 5 snippets from the OSCAR source incorrectly arguing that Barack Obama was born in Kenya.
While the creators of ROOTS employed filtering strategies targeted specifically at spam and machine-generated content \cite{laurencon2022the}, developing filters for such content is a never-ending arms race with its producers, and the only way to keep improving them is to look at the data---which our tool enables. 

\paragraph{Studying representation of dialects and social groups.}
When LLM-based systems are deployed, the implicit assumption is often that they are general-purpose and can serve all of its potential users equally well. But there is no such thing as a ``neutral'', one-size-fits-all corpus \cite{Rogers_2021_Changing_World_by_Changing_Data}. An obvious issue is dialects, and in case of multilingual models like BLOOM another obvious problem is language imbalance. Besides that, the training data may not equally represent the topics and sources associated with different demographic groups, and hence the LLM would likely not cater to them equally well. \citet{BenderGebruEtAl_2021_On_Dangers_of_Stochastic_Parrots_Can_Language_Models_Be_Too_Big} cite the example of GPT-2: the filter for its sources was that they were shared on Reddit, which overrepresents the interests of the typical Reddit user (of whom in the US 67\% are men, and 64\% are 18-29 y.o.)

Training data that is then likely to reinforce social stereotypes harmful to marginalized populations. For example, GPT-3 has been shown to over-associate Muslims with violence \cite{10.1145/3461702.3462624}. In particular, prompting the model to continue ``\textit{Two Muslims walked into...}'' tends to lead to mentions of terrorism or assault. BLOOM is not free from these biases: we sampled 10~completions and found 4 that mentioned guns or death (compared to 66\% reported for GPT-3). Exact search for ``\textit{Two Muslims Walked into...}'' returned examples of papers studying this very phenomenon, but a search for just ``\textit{Two Muslims}'' shows that many passages in OSCAR mention violence or terrorism, whereas mentions in Semantic Scholar, pseudo-crawled websites, and Wikipedia are more varied.

\paragraph{Detecting the presence of specific information.}
Where the suitability of a model to a given application depends on it being up-to-date with the latest events, or knowledge about a given fact, a tool like ours can help to quickly find out if the model even theoretically could ``learn'' a given fact. For instance, ROOTS contains 231 references to the \textit{death of Queen Elizabeth}, but they refer to the death Elizabeth~I in 1603 and not to the recent passing of Elizabeth~II in 2022.

\paragraph{Detecting plagiarism/memorization.}
Generative LLMs can memorize part of their training sets and repeat it verbatim in their outputs. We can probe an LLM to elicit candidates for data memorization~\cite{CarliniTramerEtAl_2020_Extracting_Training_Data_from_Large_Language_Models}, and the \rst can help in different ways:
\begin{itemize*}
    \item By conditioning model probing on actual training data, so that we can more easily check whether such data has been memorized;
    \item By providing the ground truth to verify that model output was part of the training data;
    \item By providing the ground truth to verify that model did have a chance to memorize something that it should have memorized;
    \item By providing match counts to identify which data was more likely to be memorized (since the number of copies in the training data influences memorization \cite{KandpalWallaceEtAl_2022_Deduplicating_Training_Data_Mitigates_Privacy_Risks_in_Language_Models}).
\end{itemize*}
For example, BLOOM correctly completes Prince Hamlet's \textit{To be or not to be} soliloquy---both using greedy decoding and nucleus sampling---but not the less popular Shakespeare quote \textit{I am in this earthly world, where to do harm... is often laudable, to do good sometime accounted dangerous folly.} With our tool we verified that BLOOM had access to at least 7 sources for the \textit{Macbeth} quote (vs at least 47 for \textit{Hamlet}), but did not ``learn'' it.

\paragraph{Verifying originality.} An important question about generative AI models is to what extent their output -- that is not a verbatim copy of training data -- can be considered original. Consider the above quote from Macbeth, which BLOOM completed for us as follows: ``\textit{I am in this earthly world, where to do harm... is to do good, and to do good is to do harm.}'' With our tool, we could easily verify that the suggested completion does not exist in the corpus verbatim. However, there are dozens of contexts where the concepts of ``good'' and ``harm'' are mentioned close to each other (esp. in the phrase ``do more harm than good''), so they were the likely indirect sources for this completion. To what degree that completion can be considered new, original text is a key question for the current discussions on plagiarism in AI writing assistants and the legal status of their output.

\paragraph{Non-existing facts.} When the same associative mechanism generates factoid text, the model may ``hallucinate'' events that never occurred---or at least, there was no evidence on which the model could draw. This, too, becomes easy to verify with our tool. BLOOM completed the prompt \textit{``When was the Golden Gate Bridge transported for the second time across Egypt?''} \cite{Hofstadter_2022_Artificial_neural_networks_today_are_not_conscious_according_to_Douglas_Hofstadter} with ``\textit{The first time was in the late 19th century, when the bridge was transported from San Francisco to Cairo}''. Of course, this ``fact'' is untrue, and was not mentioned in the corpus. But we could not even find mentions of anything else transported from San Francisco to Cairo. How exactly LLMs come up with such generations is an interesting research problem, for which tools like ours could be useful.

\paragraph{Enabling data removal requests.} The authors of texts that were included in web crawls could use such a tool to identify that fact and request the removal of their texts. For ROOTS, the data governance structure set up for Big Science workshop operated only for its duration, but should there be any future work relying on the same data hosts and agreements, the flagged data collected through our tool can be used to honor the removal requests.

\paragraph{Benchmark data contamination.} To interpret benchmark results, we need to know whether they reflect training data memorization or generalization. One approach is for the model authors to specifically plan for the evaluation benchmarks prior to training, and try to exclude the benchmark data \cite{NEURIPS2020_1457c0d6}, but this limits the options for external evaluation. Our tool enables sampled checks of benchmark data, and was already successfully used to find\footnote{\url{https://twitter.com/WilliamBarrHeld/status/1586090252946448384}} that BLOOM should not be evaluated on XNLI \cite{conneau-etal-2018-xnli}.

\paragraph{Language contamination.}
According to \citet{laurencon2022the}, ROOTS contains data in 46 languages. But this is clearly not the full story. For example, neither Danish nor Ukrainian are listed, but we found examples in these languages (stackexchange, OSCAR, parsed academic pdf data). The tool can thus be useful for investigating the transfer to ``unseen'' languages in multilingual evaluation.

\paragraph{Word sense disambiguation.}
Since the \rst provides context paragraphs, it can be used to check in what sense a word was used in the training data. For example, the acronym LLM in ROOTS is used as ``large language model'' in the parsed academic article data, but in OSCAR it means predominantly ``limited liability company'' or ``Legum Magister''. If future work extends our approach to providing search results through API, then quantitative research would also be possible with techniques like context clustering and classification.

\paragraph{Pre-processing issues.}
By searching for phrases occurring in different parts of the same document, it is possible to verify that the entire document made it through the pre-processing pipeline -- which is useful for improving it. For example, we found a news article in OSCAR, the initial paragraphs of which are missing from ROOTS.
\section{Limitations and Future Work}
A major limitation of this work is that to mitigate possible issues on the data governance side, we can only provide short snippets of the indexed texts, as is typical of web search engines. We strive to provide links to the original text sources, but this metadata is not consistently available in ROOTS.

Implementation-wise, the current version of exact search is exact down to capitalization and punctuation, and fuzzy search can be noticeably slower. These issues will be addressed in future versions.

The current tool is heavily influenced by the UX of search engines, and its core functionality is similar. In future we intend to review classic corpus analysis tools for ideas of different presentation modes, such as concordance and word sketches. We would like to add more quantitative information, e.g. term frequency information, number of hits, and co-occurrence statistics. Community feedback and suggestions are welcome in the \href{https://huggingface.co/spaces/bigscience-data/roots-search/discussions}{Community tab} of the demo. We are also pursuing a spin-off collaboration with Pyserini to make large scale indexing and hosting of textual data even more seamless.

\section{Acknowledgements}
We thank the Pyserini team---Odunayo Ogundepo, Xinyu Zhang, Akintunde Oladipo and Jimmy Lin, for their indexing insights. Big thanks to the Gradio team, especially Pete Allen, Abubakar Abid and Freddy Boulton for their support on the frontend side, and to the Hugging Face infra team for answering questions regarding hosting the tool. We thank Carlos Muñoz Ferrandis and Meg Mitchell for valuable discussions.

\section{Impact Statement}
Our tool aims to improve the current state of documentation search for large corpora of web-scraped text, starting with the ROOTS corpus. However, it also comes withe ethical considerations: for instance, it can also inadvertently display sensitive information such as PII and harmful content, and help malicious actors find information about a given topic from multiple sources (which is more difficult given only the raw text of the corpus). We are aware of these limitations, and have taken precautions to compensate for them, such as the PII redaction measures we present in Figure~\ref{fig:pii_leakage}. We also present only a snippet of the raw text, which means that for accessing the full documents, users much sign up to be a part of the Big Science organization on the Hugging Face Hub, which also reduces the amount of information that potentially malicious anonymous users can access.

\bibliographystyle{acl_natbib}
\bibliography{biblio}

\end{document}